# One-Step Abductive Multi-Target Learning with Diverse Noisy Label Samples

Yongquan Yang (remy_yang@foxmail.com)

**Abstract**

One-step abductive multi-target learning (OSAMTL) was proposed to handle complex noisy labels. In this paper, giving definition of diverse noisy label samples (DNLS), we propose one-step abductive multi-target learning with DNLS (OSAMTL-DNLS) to expand the methodology of original OSAMTL to better handle complex noisy labels.

## 1 Introduction

One-step abductive multi-target learning (OSAMTL) [1] was proposed to alleviate the situation where it is often difficult or even impossible for experts to manually achieve the accurate ground-truth labels, which leads to labels with complex noisy for a specific learning task. With a H. pylori segmentation task of medical histopathology whole slide images [1,2], OSAMTL has been shown to possess significant potentials in handling complex noisy labels, using logical rationality evaluations based on logical assessment formula (LAF) [1,3]. However, OSAMTL is not suitable for the situation of learning with diverse noisy label samples. In this paper, we aim to address this issue. Firstly, we give definition of diverse noisy label samples (DNLS). Secondly, based on the given definition of DNLS, we propose one-step abductive multi-target learning with DNLS (OSAMTL-DNLS). Finally, we provide analyses of OSAMTL-DNLS compared with the original OSAMTL.

## 2 Definition of diverse noisy label samples

Diverse noisy label samples ($DNLS$) are related to a single noisy sample ($NS$). Usually, a $NS$ consists of an instance sample ($IS$) and a corresponding noisy label sample ($NLS$). An instance sample contains a number of instances ($I$) and a corresponding noisy label sample contains a number of noisy labels ($NL$). Formally, a $NS$ can be denoted by

$$NS = \{IS, NLS\} = \{\{I_1, \cdots, I_n\}, \{NL_1, \cdots, NL_n\}\},$$

where $n$ is the number of instances or noisy labels corresponding to $NS$.

We define diverse noisy label samples ($DNLS$) for the situation, where multiple noisy label samples are assigned to the $IS$ of a single $NS$ and diversity exists between any two non-repeating noisy label samples. The diversity of two noisy label samples ($DivNLS_{a,b}$) can be evaluated by the difference between the two noisy label samples. Formally, the diversity of two noisy label samples can be denoted by

$$DivNLS_{a,b} = Differenciate\,(NLS_a, NLS_b).$$

For simplicity, we assume $DivNLS_{a,b} \in [0,1]$, where 1 signifies diversity exists between $NLS_a$ and $NLS_b$ while 0 indicates the opposite.

As a result, $DNLS$ can be formally denoted by

$$DNLS = \{NLS_1, \cdots, NLS_d\}$$

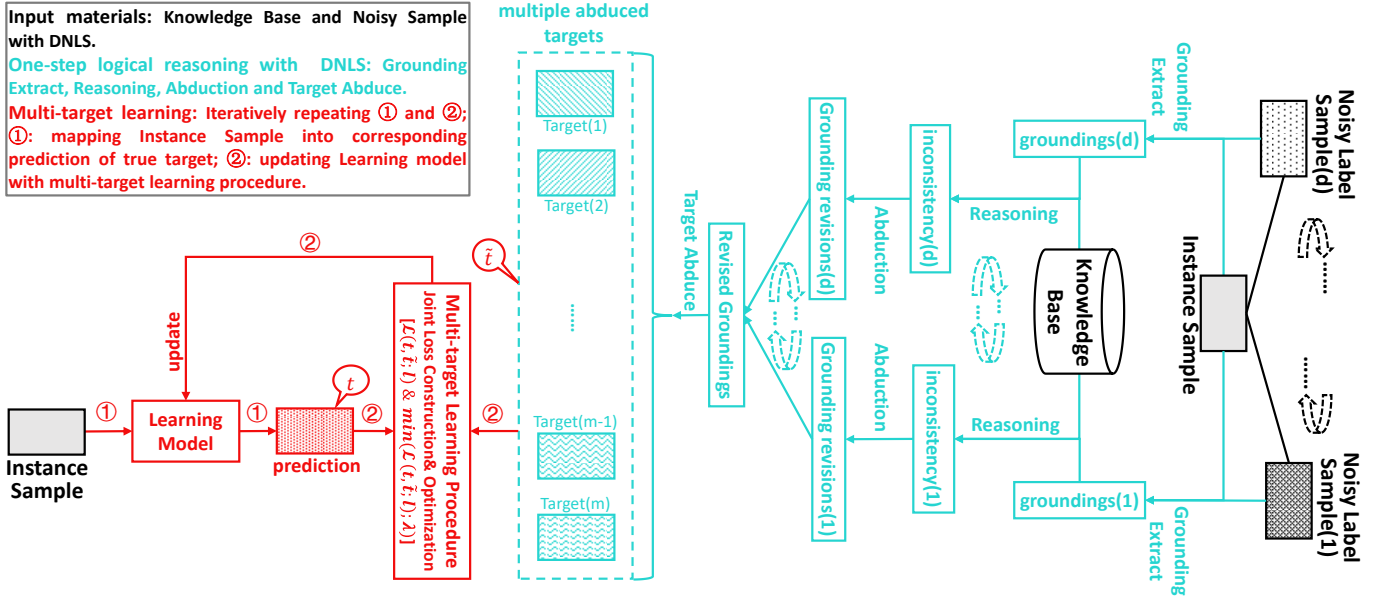

Fig. 1. The outline for the methodology of OSAMTL-DNLS. For simplicity of elaborating the methodology of OSAMTL-DNLS, we assume that the instance sample ($IS$) and each noisy label sample ($NLS$) of the diverse noisy label samples ($DNLS$) only have one instance and one noisy label respectively. This simplified elaboration can be deduced to the situation where the instance sample ($IS$) and each noisy sample ($NLS$) of the diverse noisy label samples ($DNLS$) have a set of instances and a set of noisy labels respectively.

$$= \{\{NL_{1,1}, \cdots, NL_{1,n}\}, \cdots, \{NL_{d,1}, \cdots, NL_{d,n}\}\}$$
$$s.t. \quad \forall a, \forall b \in \{1, \cdots, d\} \text{ and } a \neq b, \exists \, DivNLS_{a,b} = 1.$$

And a $NS$ with $DNLS$ can be denoted by
$$NS_{DNLS} = \{IS, DNLS\}.$$

## 3 One-step abductive multi-target learning with diverse noisy label samples

With the given definition of diverse noisy label samples (DNLS), we propose one-step abductive multi-target learning with diverse noisy label samples (OSAMTL-DNLS). OSAMTL-DNLS constitutes of three components, including input materials, one-step logical reasoning and multi-target learning. The outline for the methodology of OSAMTL-DNLS is shown as Fig. 1.

### 3.1 Input materials

The input materials of OSAMTL-DNLS include a given $NS$ with $DNLS$ and a knowledge base ($KB$) containing a list of domain knowledge about the true target of a specific task. Referring to the formulations of a $NS$ with $DNLS$ assigned presented in Section 3.1, the input materials of OSAMTL can be more specifically denoted as follows

$$KB = \{K_1, \cdots, K_b\}; NS_{DNLS} = \{IS, DNLS\}$$
$$IS = \{I_1, \cdots, I_n\},$$
$$DNLS = \{NLS_1, \cdots, NLS_d\} = \{\{NL_{1,1}, \cdots, NL_{1,n}\}, \cdots, \{NL_{d,1}, \cdots, NL_{d,n}\}\}.$$

## 3.2 One-step logical reasoning with DNLS

With the given input materials, the one-step logical reasoning procedure of OSAMTL-DNLS, which consists of four substeps, abduces multiple targets containing information consistent with the domain knowledge about the true target of a specific task.

The substep one extracts a list of groundings from the given $NS_{DNLS}$ that can describe the logical facts contained in $DNLS$. Formally, this grounding extract ($GE$) step can be expressed as

$$G = GE(\{IS, DNLS\}; p^{GE}) = \{GE(\{IS, NLS_1\}; p^{GE_1}), \cdots, GE(\{IS, NLS_d\}; p^{GE_d})\}$$
$$= \{G_1, \cdots, G_d\} = \{\{G_{1,1}, \cdots, G_{1,r_1}\}, \cdots, \{G_{d,1}, \cdots, G_{d,r_d}\}\}. \quad (1)$$

Via logical reasoning, the substep two estimates the inconsistencies between the extracted groundings $G$ and the domain knowledge in the knowledge base $KB$. Formally, this reasoning ($R$) step can be expressed as

$$IC = R(G, KB; p^R) = \{R(G_1, KB; p^{R_1}), \cdots, R(G_d, KB; p^{R_d})\}$$
$$= \{IC_1, \cdots, IC_d\} = \{\{IC_{1,1}, \cdots, IC_{1,i_1}\}, \cdots, \{IC_{d,1}, \cdots, IC_{d,i_d}\}\}. \quad (2)$$

The substep three revises the groundings of the given $NS_{DNLS}$ by logical abduction based on reducing the estimated inconsistency $Incon$. Formally, this logical abduction ($LA$) step can be expressed as

$$RG = LA(\{IC, G\}; p^{LA}) = \{LA(\{IC_1, G_1\}; p^{LA_1}), \cdots, LA(\{IC_d, G_d\}; p^{LA_d})\}$$
$$= \{\{GR_{1,1}, \cdots, GR_{1,z_1}\}, \cdots, \{GR_{d,1}, \cdots, GR_{d,z_d}\}\}$$
$$= \{RG_1(GR_{1,1}), \cdots, RG_s(GR_{d,z_d})\} \quad s.t. \quad s = \sum_{i=1}^{d} z_i. \quad (3)$$

Finally, the substep four leverages the revised groundings $RG$ and the instance samples in the given noisy samples to abduce multiple targets containing information consistent with our domain knowledge about the true target. Formally, this target abduce ($TA$) step can be expressed as

$$\tilde{t} = TA(\{RG\}; p^{TA})$$
$$= \begin{cases} TA(\{RG_{1,1}, \cdots, RG_{1,e_1}\}; p^{TA_1}), \cdots, \\ TA(\{RG_{m,1}, \cdots, RG_{m,e_m}\}; p^{TA_m}) \end{cases}$$
$$= \{\tilde{t}_1, \cdots, \tilde{t}_m\} \quad s.t. \quad \{RG_{*,1}, \cdots, RG_{*,e_*}\} \in RG. \quad (4)$$

In the four formulas (1)-(4), each $p^*$ denotes the hyper-parameters corresponding to the implementation of respective expression.

The formula (4) reflects that the instance sample of the given noisy sample with diverse noisy label samples ($IS$ of $NS_{DNLS}$) has corresponding $m$ number of abduced targets ($\tilde{t} = \{\tilde{t}_1, \cdots, \tilde{t}_m\}$). Referring to the situation where the instance sample of the given noisy sample with diverse noisy label samples only has one instance, the formular (4) can be deduced to imply that each instance contained in the instance sample of noisy sample with diverse noisy label samples has corresponding multiple abduced targets. With this implication, we can rewrite the multiple targets for the instance sample of noisy sample with diverse noisy label samples as

$$\tilde{t} = \{\tilde{t}_1, \cdots, \tilde{t}_n\} = \{\{\tilde{t}_{1,1}, \cdots, \tilde{t}_{1,m}\}, \cdots, \{\tilde{t}_{n,1}, \cdots, \tilde{t}_{n,m}\}\}. \quad (5)$$

## 3.3 Multi-target learning

The multi-target learning procedure of OSAMTL-DNS is carried out on the basis of a specifically constructed learning model that maps an input instance ($I_*$) into its corresponding target prediction ($t_*$), which can be expressed as

$$t_* = LM(I_*, \omega). \tag{6}$$

Here, $LM$ is short for learning model, and $\omega$ denotes the hyper-parameters corresponding to the construction of a specific learning model.

The multi-target learning procedure of OSAMTL-DNS, which constitutes of a joint loss construction and optimization, imposes the rearranged multiple targets ($\tilde{t}_*$) upon machine learning to constrain the prediction of the learning model ($t_*$). The joint loss is constructed by estimating the error between $t_*$ and $\tilde{t}_*$, which can be expressed as

$$\mathcal{L}(t_*, \tilde{t}_*; \ell) = \sum_{c=1}^{m} \alpha_c \ell(t, \tilde{t}_{*,c}) \ \ s.t. \ \ \sum_{c=1}^{m} \alpha_c = 1 \ and \ \tilde{t}_{*,c} \in \tilde{t}_*. \tag{7}$$

Here, $\ell$ denotes the hyper-parameters corresponding to the construction of the basic loss function, and $\alpha_c$ is the weight for estimating the loss between $t_*$ and an abduced target ($\tilde{t}_{*,c}$) contained in $\tilde{t}_*$. Then, the objective can be expressed as

$$\min_{t_*}(\mathcal{L}(t_*, \tilde{t}_*; \ell); \lambda). \tag{8}$$

Here, $\lambda$ denotes the hyper-parameters corresponding to the implementation of an optimization approach.

## 4 Analysis

OSAMTL-DNLS inherits properties from OSAMTL [1], including the difference of OSAMTL from abductive learning (ABL) [4] and the distinctiveness of OSAMTL from various state-of-the-art approaches that are based on pre-assumptions about noisy-labelled instances [5–12] or need premised requirements [13–15] to be carried out to handle noisy labels. OSAMTL-DNLS also inherits the essence of the multi-target learning procedure of OSAMTL, which is that the multi-garget learning procedure can enable the learning model to learn from a weighted summarization of multiple targets that contain information consistent to our prior knowledge about the true target of a specific task. For more details of these properties of OSAMTL-DNLS inherited from OSAMTL, readers can refer to [1].

OSAMTL-DNLS improves OSAMTL. The one-step logical reasoning procedure of OSAMTL-DNLS abduces multiple targets using the given noisy sample with divers noisy label samples and knowledge base, while the logical reasoning procedure of OSAMTL abduces multiple targets using the given noisy sample with one noisy sample and knowledge base. OSAMTL-DNLS is suitable to address tasks where a knowledge base and multiple noisy label samples are assigned to a noisy sample and each noisy label sample has a different noisy distribution. From this side, OSAMTL can be viewed as a subset of the methodology of OSAMTL-DNLS. Thus OSAMTL-DNLS expands the methodology of original OSAMTL to better handle complex noisy labels.

Because of its property of being able to learn with diverse noisy label samples, OSAMTL-DNLS can be leveraged to address some tasks in the field of medical analysis where the problem of low consistency always exists. Low consistency, here in the context of DNLS, can refer to that large is the difference between the noisy distributions of two diverse noisy label samples prepared by experts on a same instance sample (data set) for a medical analysis task.